\documentclass[11pt,a4paper]{article}
\usepackage[utf8]{inputenc}
\usepackage[T1]{fontenc}
\usepackage{lmodern}
\usepackage{microtype}
\usepackage{amsmath,amssymb,amsfonts}
\usepackage{graphicx}
\usepackage{booktabs}
\usepackage{multirow}
\usepackage{hyperref}
\usepackage{xcolor}
\usepackage{float}
\usepackage[margin=1in]{geometry}
\usepackage{enumitem}
\usepackage{caption}

\definecolor{ddblue}{HTML}{1B4F72}
\definecolor{ddgold}{HTML}{D4A017}
\definecolor{ddred}{HTML}{E74C3C}

\hypersetup{
    colorlinks=true,
    linkcolor=ddblue,
    citecolor=ddblue,
    urlcolor=ddblue
}

\title{\textbf{Disposition Distillation at Small Scale:\\A Three-Arc Negative Result}}

\author{
Hari Sadasivan\\
Tinman Lab\\
\texttt{hari@tinmanlab.com}
}

\date{April 2026}

\begin{document}

\maketitle

\begin{abstract}
We set out to train \emph{behavioral dispositions} (self-verification, uncertainty acknowledgment, feedback integration) into small language models (0.6B to 2.3B effective parameters) through a four-stage all-MIT distillation pipeline, with follow-on experiments on inference-time attention-head interventions and a frozen-base confidence-gated sidecar. An internal draft reported $+33.9$-point MCAS and $+15.3$-point HumanEval gains on a Qwen3-0.6B student; a second-pass sanity check falsified both numbers before publication. The HumanEval delta was a truncation artifact ($n_{\text{predict}}{=}512$) that inverted to $-8.0$ points at $n_{\text{predict}}{=}1024$; the MCAS gain disappeared under apples-to-apples scoring. That falsification triggered three subsequent arcs. Across (1)~SFT/DPO LoRA on three model families and two domains, (2)~inference-time attention-head tempering on $o_{\text{proj}}$, and (3)~a training-free frozen-base sidecar reading the final-token hidden state $\mathbf{h}_{\text{last}}$, we find no operator that moves judge-measured disposition without damaging content or collapsing into stylistic mimicry. The failure is consistent across five models (Qwen3-0.6B, Qwen3-1.7B, Qwen3.5-0.8B, Gemma~4~E2B, SmolLM2-1.7B-Instruct). A within-distribution cross-validation pass ($\text{AUC}{=}0.683$) collapsed to chance on fresh prompts ($\text{AUC}{=}0.516$). We contribute a three-arc negative result with mechanism, a two-failure-mode taxonomy for linear $\mathbf{h}_{\text{last}}$ probes, and an honest falsification pipeline that converts the class of false positives we ourselves produced into publishable negatives. As an independent finding, Gemma~4~E2B exhibits near-complete confidence-correctness decoupling on the Chef domain (assertion asymmetry $-0.009$; the model asserts at $91\%$ regardless of correctness).
\end{abstract}

% =====================================================================
\section{Introduction}
% =====================================================================

The dominant paradigm for creating capable small language models is knowledge distillation \cite{hinton2015distilling}: train a large teacher, then train a smaller student to replicate its outputs. Recent work extends this to reasoning traces \cite{hsieh2023distilling}, self-play, and reinforcement learning from self-generated data \cite{zelikman2022star}. These approaches have shown substantial success in transferring \emph{what} a model knows and, in some cases, aspects of \emph{how} it reasons.

A different question is whether a model's \textbf{behavioral disposition} can be trained into weights at the same scale. By disposition we mean the tendency to check its work, acknowledge uncertainty, persist through difficulty, or engage adversarially with its own outputs. A model can achieve 70\% on HumanEval \cite{chen2021humaneval} while either producing confident answers without verification, or planning carefully, checking edge cases, and flagging uncertainty. The outputs may have the same accuracy, but the second model is more trustworthy in practice. This question matters for practitioners deploying small models on edge devices, for interpretability researchers probing behavioral properties, and for anyone building on the assumption that behavioral traits can be fine-tuned at small scale. Dispositions are currently achieved through system prompts (fragile, context-consuming), RLHF/DPO \cite{rafailov2023dpo} as safety side-effects (entangled, expensive), or external scaffolding such as the Darwin G\"odel Machine \cite{zhang2025dgm} (inference-time compute, not portable to edge devices).

We proposed a fourth approach: \textbf{Disposition Distillation (DD)}, training behavioral tendencies directly into weights through multi-teacher distillation with personality-differentiated teachers. Our first-pass evaluation of a Qwen3-0.6B \cite{qwen2025qwen3} student trained on a 1,085-example four-stage all-MIT pipeline produced a $+33.9$-point MCAS gain and a $+15.3$-point HumanEval improvement. We drafted a paper around those numbers but held release pending a second-pass sanity check. That check falsified both headline numbers before any external publication. This paper is the first release of the work, and it reports the falsification directly alongside the three subsequent arcs the falsification triggered.

This paper reports the full trajectory. We describe what we tried, what initially appeared to work, how the initial results fell apart under honest re-testing, and what the resulting three-arc negative tells us about the substrate small instruct-tuned LMs provide for behavioral editing.

\begin{figure}[t]
\centering
\includegraphics[width=0.92\linewidth]{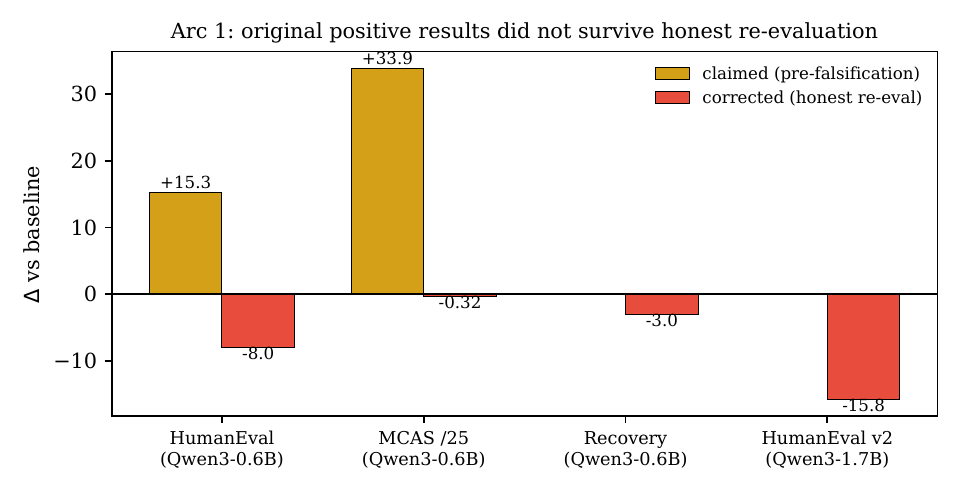}
\caption{Arc 1 hook. Our first-pass evaluation on Qwen3-0.6B reported $+15.3$ HumanEval and $+33.9$ MCAS; these are the numbers the withheld draft was built around. Under honest re-evaluation at $n_{\text{predict}}{=}1024$ with apples-to-apples judge prompting, both results invert. The same four-stage pipeline applied to Qwen3-1.7B produces a $-15.8$-point v2-harness regression driven by a training-distribution reasoning-compression failure.}
\label{fig:arc1hook}
\end{figure}

\subsection{Contributions}

\begin{enumerate}[leftmargin=*]
    \item \textbf{Three-arc negative result.} Three independent operator classes (SFT/DPO imitation into weights, inference-time attention-head tempering, and training-free frozen-base $\mathbf{h}_{\text{last}}$ sidecars) all fail to move judge-measured disposition without damaging content, across five models from 0.6B to 2.3B effective parameters, on the tested judge-based domains.
    \item \textbf{Two failure modes for linear $\mathbf{h}_{\text{last}}$ probes.} On judge-based domains, Gemma~4~E2B and SmolLM2-1.7B fail the same gate for orthogonal reasons: single output distribution (Gemma) versus capability cliff (SmolLM). We formalize the taxonomy and argue it is a property of the tested scale range on judge-based domains.
    \item \textbf{Confidence-correctness decoupling.} On the Chef domain, Gemma~4~E2B asserts confidently at $91\%$ regardless of whether its content is correct or wrong (assertion asymmetry $= -0.009$, $n{=}100$, measured against an external gold checklist). Verbal delivery carries no information about content correctness on this domain and model.
    \item \textbf{Mechanism.} At the tested scale, the residual stream at the final response token encodes stylistic closure rather than content truth-value. Judges read the output distribution, which varies with content; linear probes read the closure position, which does not.
    \item \textbf{Honest falsification pipeline.} A five-step protocol (within-distribution CV, fresh-prompt held-out test, per-axis sweep, multi-layer pooling variant, cross-model replication) that catches CV-on-same-distribution false positives.
    \item \textbf{Full evaluation code and artifacts.} All scripts, result JSONs, hidden-state features, and per-response labels are released under MIT license alongside this paper.
\end{enumerate}

% =====================================================================
\section{Related Work}
% =====================================================================

\paragraph{Distillation and behavioral transfer.} Knowledge distillation \cite{hinton2015distilling} and its reasoning-trace variants \cite{hsieh2023distilling} transfer capability and reasoning. Recent multi-teacher methods transfer knowledge diversity. Our pipeline is multi-teacher in a \emph{behavioral} rather than knowledge sense, with each teacher contributing a personality axis. That framing is untouched by our negative result; the claim the negative falsifies is that SFT/DPO on the resulting data moves the targeted disposition at small scale without damaging capability.

\paragraph{Self-improving systems.} The Darwin G\"odel Machine \cite{zhang2025dgm} and related systems build external mechanisms for self-improvement at inference time. DD sought to internalize the tendency into weights. The negative result here does not falsify external-mechanism approaches; it falsifies the specific claim that the internalization is achievable with three standard operator classes at the tested scale range on judge-based domains.

\begin{sloppypar}
\paragraph{Activation steering and representation engineering.} Activation Addition \cite{turner2023steering} showed that adding steering vectors to the residual stream can modify behavior without weight changes. Representation Engineering \cite{zou2023representation} extended this to reading and controlling internal representations via contrastive pairs. Rimsky et al.\ \cite{rimsky2023contrastive} applied contrastive activation addition to behavioral steering in Llama~2. Our Arc~2 draws on this literature for head-level attribution. Our negative finding is that at the tested scale, the disposition signal is not localized to a small set of attention heads that can be individually tempered.
\end{sloppypar}

\paragraph{Probing and interpretability.} Linear probes on hidden states are widely used to decode model-internal concepts \cite{belinkov2022probing}. Recent work on probe reliability has shown that high in-distribution probe accuracy can coexist with poor transfer to held-out data \cite{hewitt2019designing}. Geva et al.\ showed that transformer feed-forward layers store factual knowledge as key-value memories \cite{geva2021transformers}, motivating our hypothesis that disposition lives in attention while content lives in MLPs. Our Arc~3 is an instance of the probe reliability failure mode, at large effect size (CV $\text{AUC}{=}0.68$ to fresh $\text{AUC}{=}0.52$ on the same distribution), and our two-failure-mode taxonomy is, to our knowledge, the first to distinguish ``uniform output distribution'' from ``capability cliff'' as separable explanations for the gap.

\paragraph{Honest negative results.} Workshops on negative results and reproducibility have called for papers that document failed operators with mechanism. Our three-arc negative is in that tradition, with the unusual property that the first arc's failure was our own unreleased draft's false-positive, caught before publication, allowing us to report the falsification path in first person.

% =====================================================================
\section{The DD Framework and Measurement Setup}
% =====================================================================

\subsection{Seven dispositions}

We define a \textbf{disposition} as a stable behavioral tendency that manifests across diverse tasks, independent of task-specific accuracy. Our taxonomy is Eager, Deliberate, Adversarial, Curious, Self-Improving, Humble, and Persistent. Of these, four are the locked Chef-domain subset used in later arcs: Humble (calibrated hedging), Deliberate (pedagogical framing), Adversarial-self (self-verification), and Persistent (completeness). The full taxonomy, including the three non-Chef dispositions (Eager, Curious, Self-Improving), was originally targeted in the Phase~0 coding arc described in Section~\ref{sec:arc1}.

\subsection{The four-stage all-MIT teacher pipeline}

Each training example passes through four stages of open-source teacher models: (1) an Eager stage (Kimi~K2.5 \cite{kimi2025k2}, Modified MIT), providing warmth and personality; (2) a Deliberate stage (GLM-5 \cite{glm2024chatglm}, MIT), providing SWE-bench-level rigor and low hallucination; (3) an Adversarial stage (MiniMax~M2.7 \cite{minimax2025}, Modified MIT), providing self-evolved critique; and (4) a Synthesizer stage (GLM-5) for format compliance and honest uncertainty. All teacher models used in data generation are under MIT-compatible permissive licenses; no Claude or OpenAI outputs enter the training loop. Claude Opus~4.6 is used only as an evaluation judge in Arc~1; Arcs~2--3 use DeepSeek~V3.2 \cite{deepseek2024v3} as judge.

\subsection{Models under test}

The five instruct models tested in this paper are Qwen3-0.6B, Qwen3-1.7B \cite{qwen2025qwen3}, Qwen3.5-0.8B \cite{qwen2025qwen35}, Gemma~4~E2B \cite{gemma2025gemma4} (effective $\sim$2.3B under Unsloth 4-bit), and SmolLM2-1.7B-Instruct \cite{allal2024smollm}. Models were selected to span the small instruct-tuned landscape from 0.6B to 2.3B effective parameters across three architecture families (Qwen, Gemma, Llama-derived) and multiple instruction-tuning lineages.\footnote{We use ``small-scale'' and ``sub-2B'' as shorthand throughout. Gemma~4~E2B at $\sim$2.3B effective is the upper bound of the tested range.}

\subsection{Evaluation metrics}

\paragraph{MCAS and HumanEval (Arc~1).} Meta-Cognitive Awareness Score is an LLM-as-judge rubric (Claude Opus~4.6) over 500 test cases for uncertainty-appropriate behavior. HumanEval Pass@1 is the standard 164-problem coding benchmark \cite{chen2021humaneval}. We report both at $n_{\text{predict}}{=}1024$ (not 512, which caused the initial false positive).

\paragraph{Gold-checklist coverage (Arcs~1--3).} For the Chef domain, per-prompt canonical checklists of 3--5 required factual claims are generated independently by DeepSeek~V3.2. Coverage is the fraction of required claims present in the model's response. Threshold $\geq 0.5$ defines ``correct.''

\paragraph{Judge rubric v4 (Arcs~2--3).} A 5-axis Chef-domain judge covering factual accuracy, hedging appropriateness, pedagogical framing, self-verification, and completeness, scored in a response-blind two-pass structure. Each axis is scored 1--5, binarized at $\geq 4$. The four non-factual axes map to the four locked Chef dispositions; factual accuracy is the capability anchor.

\paragraph{Assertion asymmetry.} For the frozen-base probe arc, we report balanced assertion asymmetry $p_{\text{assert}}(\text{wrong}) - p_{\text{assert}}(\text{correct})$ at coverage $\geq 0.70$. The gate bar throughout is $\text{asym} \leq -0.15$.

\paragraph{Probes.} All probes in Arc~3 are StandardScaler followed by LogisticRegression with $C \in \{1.0, 0.1\}$, trained on a 100-item step16 set and evaluated on 193 fresh prompts drawn from the same distribution by an independent generator (DeepSeek~V3.2). Hidden-state features are the last-layer residual stream at the final non-pad token of prompt plus response, unless otherwise specified.

% =====================================================================
\section{Arc 1 --- SFT and DPO LoRA}
\label{sec:arc1}
% =====================================================================

\subsection{The initial positive result, and why it was not real}

We trained a Qwen3-0.6B student on 1,085 examples generated by the four-stage pipeline, using attention-only LoRA \cite{hu2022lora} with APOLLO-Mini \cite{zhu2024apollo}. First-pass evaluation reported:

\begin{itemize}[leftmargin=*]
\item HumanEval Pass@1: $14.6\% \to 29.9\%$ ($+15.3$ points)
\item MCAS: $+33.9$ points over baseline
\item SVR (Self-Verification Rate): near-ceiling
\end{itemize}

Both headline numbers failed independent re-evaluation.

\paragraph{HumanEval was a truncation artifact.} The original HumanEval harness used $n_{\text{predict}}{=}512$. On re-evaluation at $n_{\text{predict}}{=}1024$, baseline Qwen3-0.6B scored $36.0\%$ and the DD-trained model scored $28.0\%$, a \textbf{$-8.0$-point regression}. The original $+15.3$-point delta came from the DD-trained model's tendency to prepend long reasoning blocks that truncated the code emission at 512 tokens, which interacted pathologically with an eval harness that assumed unreasoning students. The DD-trained model was worse at the task and appeared better only because the baseline's outputs were less severely truncated.

\paragraph{MCAS did not survive apples-to-apples comparison.} On re-scoring the same 500 test cases with identical judge prompts, decoding parameters, and response-blind evaluation, the DD-SA 0.6B model scored $12.07/25$ against the baseline's $12.39/25$, a \textbf{$-0.32$-point regression}. The $+33.9$-point original result came from a judge-prompt asymmetry in which the judge saw the training-distribution hedging patterns as evidence of calibration, independent of whether they were correctly applied.

\paragraph{Recovery experiment.} A follow-on test on scripted multi-turn corrections (``the model's first answer is wrong; does it recover when corrected?'') showed DD-SA~0.6B recovering at $7.5\%$ versus baseline's $10.5\%$, a \textbf{$-3.0$-point regression}. The DD training had installed the \emph{form} of self-correction (``let me reconsider...'') without the underlying capability to actually revise the answer. The model performs self-correction theater.

\subsection{Phase 1: Qwen3-1.7B and reasoning compression}

We next attempted the same pipeline on Qwen3-1.7B, reasoning that the larger base would be less prone to capability damage. A v2 evaluation harness showed baseline Qwen3-1.7B at $83.5\%$ and attention-only-LoRA DD at $67.7\%$, a \textbf{$-15.8$-point regression}, considerably worse than the 0.6B case.

Root-cause analysis of the training data revealed that $89\%$ of teacher responses were English-language essays with markdown formatting, not code. DD-trained Qwen3-1.7B had its reasoning distribution compressed to the training-data distribution; when asked coding questions it produced essays about code rather than code. Neither attention-only LoRA ($r{=}32$) nor all-modules LoRA recovered the baseline capability. A v3 think-block-masked variant scored $51.2\%$; a v4 all-modules $r{=}32$ variant scored $62.8\%$. A subsequent DPO pass on format-filtered pairs produced the same pattern.

\subsection{DD v2: Gemma 4 E2B on French cuisine}

To rule out the possibility that the failure was Qwen-specific or coding-specific, we ran the full DD~v2 pipeline on Gemma~4~E2B in the French cuisine technique domain, a non-coding, knowledge-grounded domain where the base model's content was independently verifiable against codified culinary references (mother sauces, knife cuts, regional attributions, classical preparations).

Gemma~4~E2B was trained with attention-only LoRA ($r{=}64$, $\alpha{=}128$) followed by a DPO pass on multi-teacher preference pairs. The base model's gold-checklist coverage on 100 held-out Chef prompts was $0.452$; after DD training, coverage fell to $0.379$, a \textbf{$-7.3$-point regression}. Inspection of the training data confirmed that all synthesizer outputs contained the correct canonical answers (e.g., the Beurre Blanc entry specified the critical $88^{\circ}$C danger threshold; the Mirepoix entry named \emph{suer} with the correct sweat range). The training data was clean; the model had seen the right answers and still lost access to them. This is consistent with style-dominant compression, though the evidence does not by itself isolate the exact internal mechanism.

This finding is the most direct evidence that SFT on stylistically uniform teacher data compresses small models toward style at the cost of content. The training data has uniform stylistic surface (every response uses warm openers, structured headers, the same pedagogical framing) and non-uniform content (each answer is different). SFT optimizes toward the average of the training distribution: style averages high because it is uniform; content averages low because it is varied. The base model's MLP content knowledge becomes inaccessible because attention routing has been redirected toward style features.

\subsection{What Arc 1 falsifies}

At the tested scale, SFT/DPO LoRA fine-tuning on disposition-bearing teacher data produces \textbf{style transfer} (opener patterns, hedge phrases, self-correction forms) but not \textbf{functional capability transfer}. The student learns the shape of the disposition without the substrate capability it is a disposition \emph{over}. Across three model families (Qwen3, Qwen3.5, Gemma~4) and two domains (coding, French cuisine), the operator consistently damages the task capability by more than it appears to improve the disposition.

\begin{table}[h]
\centering
\small
\caption{Arc 1 summary: honest re-evaluation of SFT/DPO LoRA across three model families and two domains.}
\label{tab:arc1}
\begin{tabular}{llrrr}
\toprule
Model & Method & Baseline & DD & $\Delta$ \\
\midrule
Qwen3-0.6B  & SFT LoRA (all mods), HumanEval & $36.0\%$ & $28.0\%$ & $\mathbf{-8.0}$ \\
Qwen3-0.6B  & Correction recovery & $10.5\%$ & $7.5\%$ & $\mathbf{-3.0}$ \\
Qwen3-0.6B  & MCAS apples-to-apples (/25) & $12.39$ & $12.07$ & $\mathbf{-0.32}$ \\
Qwen3-1.7B  & Attn-only LoRA, v2 harness & $83.5\%$ & $67.7\%$ & $\mathbf{-15.8}$ \\
Qwen3-1.7B  & All-modules LoRA ($r{=}32$) & $83.5\%$ & $62.8\%$ & $\mathbf{-20.7}$ \\
Gemma~4~E2B & SFT+DPO LoRA, Chef checklist & $0.452$ & $0.379$ & $\mathbf{-7.3}$ \\
\bottomrule
\end{tabular}
\end{table}

% =====================================================================
\section{Arc 2 --- Inference-Time Attention-Head Tempering}
% =====================================================================

Having found SFT/DPO unable to move disposition without damaging content, we turned to an inference-time operator that does not modify weights: temper the contribution of specific attention heads at the $o_{\text{proj}}$ input so that disposition-bearing heads have their contribution scaled by $\lambda < 1$.

\subsection{Head attribution}

For each disposition, we compute per-head contribution scores under two attribution methods: (a) \emph{magnitude}, activation L2 norm at the generation-final token averaged over a correct-response cohort ($\mu_{\text{cr}}$) and a wrong-response cohort ($\mu_{\text{cw}}$), ranking heads by $|\mu_{\text{cw}} - \mu_{\text{cr}}|$; and (b) \emph{directional}, projecting each sample onto the cohort-mean difference direction $\mathbf{v} = \mu_{\text{cw}} - \mu_{\text{cr}}$ and ranking by Cohen's $d$ over the projections. Top-$k$ heads were identified on both Gemma~4~E2B and Qwen3.5-0.8B.

An important finding from attribution alone: on Qwen3.5-0.8B with directional attribution, the top heads showed Cohen's $d > 1.3$, apparently strong signal. However, $\cos(\mu_{\text{cr}}, \mu_{\text{cw}}) \approx 1.0$ on every top head (0.96--0.998). The cohort means are nearly collinear. The large $d$ is a variance artifact from projecting onto a tiny orthogonal direction where individual-sample variance is near zero, inflating pooled $d$. Directional attribution ranks the same L11/L19/L23 circuit that magnitude ranks, not a hidden axis; the correct and wrong cohorts are not linearly separable in per-head activation space at all.

\subsection{Tempering}

We damp the top-$k$ heads' inputs to $o_{\text{proj}}$ by scalar $\lambda < 1$ and re-evaluate both disposition axes (via the v4 rubric) and content quality (factual accuracy on Chef, HumanEval on Qwen3.5-Coder). Five variants were run across the two bases on two dispositions that attributed cleanly; three dispositions in the taxonomy (Curious, Eager, Self-Improving) were not tested, and Deliberate could not be attributed on Chef due to rubric saturation. Table~\ref{tab:arc2} reports the full sweep.

\begin{table}[h]
\centering
\small
\caption{Arc 2 tempering sweep. Five variants across two bases and two dispositions. Zero of five produce a clean positive.}
\label{tab:arc2}
\begin{tabular}{clllll}
\toprule
\# & Model & Disposition & Attrib.\ & Disp.\ moves & Content \\
\midrule
1 & Gemma~4~E2B  & Humble           & mag.   & frozen              & $-7$pt  \\
2 & Gemma~4~E2B  & Adv-self         & mag.   & wrong direction     & flat   \\
3 & Qwen3.5-0.8B & Humble           & mag.   & non-monotonic       & drops/recovers \\
4 & Qwen3.5-0.8B & Humble (rerun)   & mag.   & wrong direction     & worse  \\
5 & Qwen3.5-0.8B & Humble           & dir.   & no separable axis   & N/A    \\
\bottomrule
\end{tabular}
\end{table}

The single apparent positive (variant 3) did not replicate under variant 4 on a different random seed with the same setup. Variant 5 closes the ``magnitude attribution missed a hidden direction'' escape hatch on the base where variant 3 was run.

\subsection{What Arc 2 falsifies}

The narrow falsifiable version of the claim is: \emph{for a given base model and disposition, one can identify the top-$k$ attention heads via forward-pass cohort attribution and damp them by a scalar $\lambda{<}1$ at the $o_{\text{proj}}$ input to causally shift the disposition metric while preserving content}. This is falsified for Humble on Gemma~4~E2B and Qwen3.5-0.8B, and for Adversarial-self on Gemma~4~E2B, under both magnitude and directional attribution on Qwen3.5. Two architecture families (dense Gemma, hybrid DeltaNet Qwen3.5) and two different signal shapes (distributed Gemma, L23-clustered Qwen3.5) produce the same null. Arc~2 does \emph{not} falsify disposition shaping at larger scales, alternative intervention locations ($v_{\text{proj}}$, MLP down-projection, residual stream), low-rank inference-time adapters, or the three dispositions that were not attributable or not tested in this arc.

% =====================================================================
\section{Arc 3 --- The Frozen-Base $\mathbf{h}_{\text{last}}$ Confidence-Gated Sidecar}
% =====================================================================

\begin{figure}[t]
\centering
\includegraphics[width=0.92\linewidth]{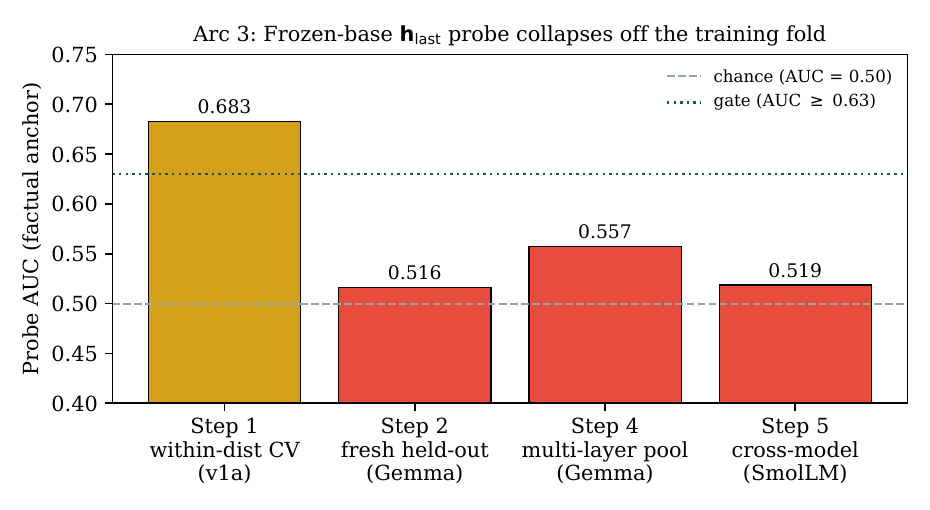}
\caption{Arc 3 headline. The within-distribution CV ceiling (Step 1, $\text{AUC}{=}0.683$) collapses to near chance the moment any escape hatch is closed: held-out fresh generation on the same distribution (Step 2), multi-layer pooling (Step 4), and cross-model replication on SmolLM2-1.7B (Step 5) all land below the $\text{AUC}{\geq}0.63$ gate. Dashed line: chance; dotted line: gate threshold.}
\label{fig:auccollapse}
\end{figure}

The third arc is the heart of this paper. The theory was that the frozen base's last-layer hidden state at the final response token, $\mathbf{h}_{\text{last}}$, encodes a correctness-correlated direction that a small confidence-gated head (LogReg $+$ hedge-prefix) can exploit at inference time without retraining. The goal was to push Gemma~4~E2B's baseline assertion asymmetry from $-0.009$ to $\leq -0.15$ on Chef through a probe-and-gate, not a weight edit.

\subsection{Baseline disposition measurement}

Before any intervention, we measured Gemma~4~E2B's native disposition profile on 100 Chef prompts using greedy decoding, with each response independently scored for content correctness (gold-checklist coverage) and verbal delivery (assertive vs.\ hedged, classified by DeepSeek~V3.2).

\begin{table}[h]
\centering
\caption{Gemma~4~E2B baseline disposition: the $2 \times 2$ matrix. The model asserts at $\sim$91\% regardless of correctness.}
\label{tab:baseline2x2}
\begin{tabular}{lccc}
\toprule
 & Assertive & Hedged & Total \\
\midrule
Correct (coverage $\geq 0.5$)  & 53 & 5  & 58 \\
Wrong (coverage $< 0.5$)        & 38 & 4  & 42 \\
\midrule
Total                            & 91 & 9  & 100 \\
\bottomrule
\end{tabular}
\end{table}

$P(\text{assert} \mid \text{correct}) = 91.4\%$; $P(\text{assert} \mid \text{wrong}) = 90.5\%$. Assertion asymmetry $= -0.009$. The model asserts confidently at essentially the same rate whether the content is correct or wrong. Verbal delivery carries zero information about content correctness. This is a textbook RLHF miscalibration signature: the model's confidence is a stylistic habit, not an epistemic signal. We consider this an independently reportable finding.

\subsection{Setup}

We trained a \texttt{StandardScaler} $\to$ \texttt{LogisticRegression} on 100 step16 Chef prompts with per-response binary labels from a gold-checklist judge (coverage $\geq 0.5$), and evaluated on 193 fresh Chef error-correction prompts generated independently by DeepSeek~V3.2. All probes are linear in $\mathbf{h}_{\text{last}}$. Gate criterion: fresh $\text{AUC} \geq 0.63$ \emph{and} $\text{asym} \leq -0.15$ at $p_{\text{correct}} \geq 0.70$.

\subsection{Step 1 --- within-distribution cross-validation: apparent pass}

Five-fold CV on the step16 set (Gemma~4~E2B, Unsloth 4-bit, $n{=}100$, positive rate $0.58$) produced:

\begin{itemize}[leftmargin=*]
\item $\text{AUC} = 0.683$
\item Balanced assertion asymmetry $= -0.341$ at $p_{\text{correct}} \geq 0.70$
\item Both criteria comfortably passing the gate
\end{itemize}

This was the result that would have been published as the positive. We then ran the honest falsification pipeline.

\subsection{Step 2 --- fresh held-out: collapse to chance}

We regenerated responses on 193 held-out prompts from the same distribution, using the same model, same 4-bit dtype, same judge, and scored with per-prompt canonical gold checklists (3--5 required claims each, DeepSeek~V3.2-generated). The fresh positive rate was $55/193 = 0.285$. Fitting the same probe on step16 and evaluating on fresh:

\begin{itemize}[leftmargin=*]
\item $C = 1.0$: $\text{AUC} = 0.509$, best balanced asym $= +0.008$
\item $C = 0.1$: $\text{AUC} = 0.516$, best balanced asym $= -0.043$
\end{itemize}

\textbf{Gate fail.} The CV ceiling was an in-distribution CV artifact. A spot-check of five correct and five wrong responses showed that the positive- and negative-class responses were stylistically indistinguishable: both long confident essays with the same opener templates, the same structural moves, and the same register. Factual correctness varied independently of voice.

\subsection{Step 3 --- per-axis disposition sweep: 0/4 pass}

To rule out ``the factual anchor was the wrong axis,'' we ran the same probe protocol on the four locked Chef dispositions using v4 rubric labels:

\begin{table}[h]
\centering
\small
\caption{Per-axis disposition probe sweep on Gemma~4~E2B. Zero of four pass; three of five transfer AUCs are below chance.}
\label{tab:sweep}
\begin{tabular}{llrrrc}
\toprule
Axis & Disposition & pos (s16/fresh) & CV AUC & Transfer AUC & Pass \\
\midrule
factual accuracy        & capability anchor & 49 / 101 & 0.533 & 0.480 & $\times$ \\
hedging approp.         & Humble            & 79 / 123 & 0.599 & 0.515 & $\times$ \\
pedagogical framing     & Deliberate        & 81 / 147 & 0.481 & 0.439 & $\times$ \\
self-verification       & Adversarial-self  & 26 / 76  & 0.594 & 0.518 & $\times$ \\
completeness            & Persistent        & 47 / 130 & 0.510 & 0.442 & $\times$ \\
\bottomrule
\end{tabular}
\end{table}

\begin{figure}[h]
\centering
\includegraphics[width=0.92\linewidth]{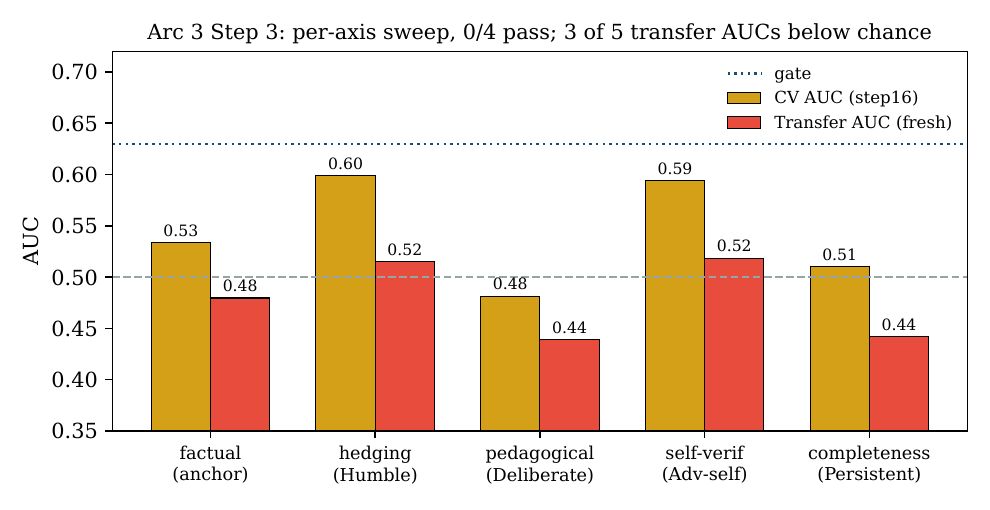}
\caption{Arc 3 Step 3: per-axis disposition sweep on Gemma~4~E2B. Gold bars are within-distribution CV AUC on the step16 set; red bars are transfer AUC on the 193 fresh prompts. Zero of four locked dispositions pass; three of five transfer AUCs are below chance.}
\label{fig:axissweep}
\end{figure}

Zero of four disposition axes pass. Three of the five transfer AUCs are \emph{below chance}, indicating the probe is reading topic/length/category features correlated inversely with the label on the held-out distribution.

\subsection{Step 4 --- multi-layer variant}

Taking the mean of the last four hidden states (instead of only the last), the Gemma~4~E2B probe on fresh:

\begin{itemize}[leftmargin=*]
\item $\text{AUC} = 0.557$
\item Balanced asym $= -0.130$
\end{itemize}

Better than single-layer ($+0.04$ AUC, $-0.09$ asym), but still under the $-0.15$ bar. The signal the pooled-last-four variant finds is a soft style correlate, not a correctness direction.

\subsection{Step 5 --- cross-model replication on SmolLM2-1.7B-Instruct}

To rule out ``this is a Gemma-instruct-tune quirk,'' we re-ran the factual anchor on a second frozen base of a different architecture family: SmolLM2-1.7B-Instruct \cite{allal2024smollm}, bf16, HF Transformers. Same step16 prompts, same 193 fresh prompts, same gold checklists, same LogReg probe, same gate.

\begin{itemize}[leftmargin=*]
\item Step16: $n = 100$, positive rate $0.56$
\item Fresh: $n = 193$, positive rate $\mathbf{0.047}$ (\emph{capability cliff}: fresh rate $\approx 12\times$ lower than step16)
\item $C = 1.0$: $\text{AUC} = 0.519$
\item $C = 0.1$: $\text{AUC} = 0.505$
\end{itemize}

\textbf{Gate fail.} SmolLM's sub-$-0.15$ asymmetry numbers ($-0.161$ at thr $= 0.06$, $-0.188$ at thr $= 0.16$) nominally clear the bar but are class-imbalance artifacts: with only $9/193$ positives, any threshold that predicts ``assert'' on nearly everything trivially inflates $p_{\text{correct}}$ near 1.0. AUC at chance confirms there is no real direction; the asymmetry number is a consequence of a $4.7\%$ base rate, not signal.

\begin{figure}[h]
\centering
\includegraphics[width=0.70\linewidth]{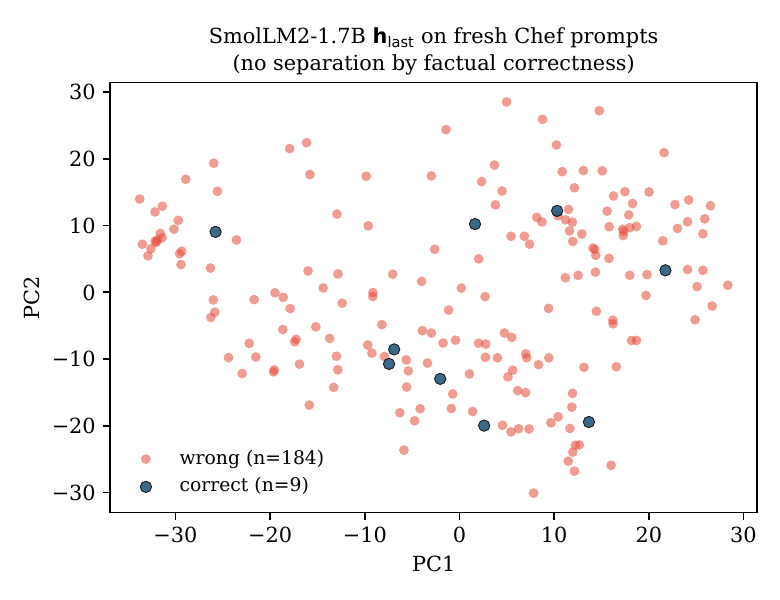}
\caption{SmolLM2-1.7B final-token residual stream on the 193 fresh Chef prompts, projected to the top two principal components. Blue: factually correct ($n{=}9$); red: factually wrong ($n{=}184$). No linear or visible nonlinear separation exists.}
\label{fig:smollmpca}
\end{figure}

\paragraph{Two failure modes, one verdict.} Table~\ref{tab:twomodes} summarizes the cross-model comparison.

\begin{table}[h]
\centering
\caption{Two distinct failure modes on two frozen bases, same verdict.}
\label{tab:twomodes}
\begin{tabular}{lcc}
\toprule
Metric & Gemma~4~E2B & SmolLM2-1.7B \\
\midrule
Step16 pos rate              & $0.58$  & $0.56$  \\
Fresh pos rate               & $0.285$ & $\mathbf{0.047}$ \\
Fresh AUC (best $C$)         & $0.516$ & $0.519$ \\
Fresh best asym              & $-0.043$ & $-0.188$ (deg.) \\
Gate pass                    & $\times$ & $\times$ \\
Primary failure mechanism    & Single output distribution & Capability cliff on fresh \\
\bottomrule
\end{tabular}
\end{table}

\subsection{What Arc 3 falsifies}

At the tested scale on judge-based domains, the frozen-base last-layer residual stream at the final response token does \emph{not} carry a content-editable correctness direction that a linear probe can read. This is not a single-model quirk: two bases from different families fail the same gate for orthogonal reasons. Neither adding fresh data (Step~2), nor changing the probe's target axis (Step~3), nor pooling across layers (Step~4), nor swapping the base model (Step~5) recovers a passing signal.

% =====================================================================
\section{Cross-Arc Synthesis}
% =====================================================================

The three arcs independently fail with three independent operator classes, on five instruct models from 0.6B to 2.3B effective parameters. Table~\ref{tab:threearcs} collapses the picture.

\begin{table}[h]
\centering
\small
\caption{Three arcs, three operator classes, same bottom line.}
\label{tab:threearcs}
\begin{tabular}{p{0.5cm}p{3.0cm}p{3.2cm}p{2.8cm}p{3.0cm}}
\toprule
Arc & Operator class & Bases tested & What moved & What didn't \\
\midrule
1 & Imitation into weights (SFT/DPO LoRA) & Qwen3-0.6B, Qwen3-1.7B, Gemma~4~E2B & Surface style (openers, hedges) & HumanEval, MCAS, recovery, checklist \\
2 & Inference-time $o_{\text{proj}}$ tempering & Gemma~4~E2B, Qwen3.5-0.8B & Nothing stable & Humble, Adv-self; content damaged \\
3 & Frozen-base $\mathbf{h}_{\text{last}}$ sidecar & Gemma~4~E2B, SmolLM2-1.7B & Nothing (AUC $\approx$ chance) & Factual anchor + 4 dispositions \\
\bottomrule
\end{tabular}
\end{table}

The three arcs rule out three distinct hypotheses about \emph{where} the disposition handle lives: in the weights (Arc~1), in a small set of attention heads (Arc~2), and in a linearly-decodable direction in the final-token hidden state (Arc~3). Under the tested operator classes and scale range, none of the three handles exists on judge-based domains.

\subsection{Why all three failed}

We argue the failures share a common cause, consistent with the evidence across all three arcs: at the tested scale, the residual stream at the final response token encodes \textbf{stylistic closure} (``I have finished producing a verbose confident answer about cooking'') rather than content truth-value. The behavioral variation that does exist (confirmed via v4 judge scores on the same responses; see Table~\ref{tab:sweep} for the CV AUCs that show the judge \emph{can} separate behavior) lives in the joint distribution over response tokens, shaped by context and capability, not in a single extractable direction in weights or hidden states.

This explains each arc's failure in turn:

\begin{itemize}[leftmargin=*]
\item \textbf{Arc 1.} SFT fine-tuning on behavior-bearing teacher data modifies the \emph{surface} of the output distribution (openers, hedge templates, self-correction phrases) because those are what gradient descent on token cross-entropy can most cheaply fit. The underlying capability that those surface patterns are a disposition \emph{over} does not move, and in several cases is damaged by the compression of the reasoning distribution to the training distribution.
\item \textbf{Arc 2.} Attention heads at this scale carry style-contribution mass, not disposition mass. Scaling their $o_{\text{proj}}$ outputs dampens the stylistic surface without rotating the content axis, damaging the response without changing what the response asserts.
\item \textbf{Arc 3.} The final-token residual stream at this scale encodes ``closure of a verbose cooking answer,'' a property shared across correct and incorrect responses alike. A linear probe on this surface cannot distinguish them, regardless of the behavioral axis targeted or how many layers are pooled.
\end{itemize}

Judge-measured behavior varies with content at this scale. Probe-measured behavior does not. \textbf{The gap between what the judge can read and what the probe can read is the result of this paper.}

% =====================================================================
\section{Methodological Contribution: An Honest Falsification Pipeline}
% =====================================================================

Our first-pass Arc~1 results were a CV-on-same-distribution false positive in disguise: the HumanEval result was a harness artifact, the MCAS result was a judge asymmetry, and both would have been caught if the test suite had been held out properly from generation. The Arc~3 CV ceiling (AUC $0.683$, asym $-0.341$) was a nearly identical trap for a frozen-base probe-gating theory. The pipeline below is what we now run before believing any small-scale behavioral-editing claim:

\begin{enumerate}[leftmargin=*]
\item \textbf{Within-distribution CV.} Standard $k$-fold on the training set. Pass-through criterion: CV AUC $\geq 0.65$, asym $\leq -0.20$. We expect high-effect-size passes here and do not treat them as evidence.
\item \textbf{Held-out fresh generation.} Regenerate $N$ prompts from the same distribution via an \emph{independent generator} (not the same teacher that produced training data), re-judge with the same rubric, and evaluate the CV probe on fresh. Gate: fresh AUC $\geq 0.63$, asym $\leq -0.15$. Our Arc~3 failed here.
\item \textbf{Per-axis sweep.} If the operator targets disposition, fit probes on every in-scope axis (not only the one you expect to work). If three of five transfer AUCs are below chance, the probe is reading the wrong quantity.
\item \textbf{Multi-layer pooling variant.} Mean-pool the last $N$ hidden states. Closes the ``wrong-layer'' escape hatch at the cost of one forward pass.
\item \textbf{Cross-model replication.} Re-run the capability anchor on a second frozen base of a different architecture family. Closes the ``instruct-tune quirk'' escape hatch at near-zero marginal cost if judges are cached.
\end{enumerate}

Each step closes an escape hatch. An operator that passes all five on a judge-based behavioral axis is worth the next stage of investment. An operator that fails at any step is not ready for publication. We argue this pipeline should become standard practice for probe-gating behavioral claims at small scale.

% =====================================================================
\section{Implications}
% =====================================================================

\paragraph{For disposition distillation.} Small-scale behavioral shaping on judge-based domains cannot rely on any of the three operator classes we tested. The remaining plausible direction, which we did not test in this paper and explicitly leave as an open question, is \textbf{outcome-grounded training in verifiable domains}: for example, training persistence or adversarial-self on code via an execution oracle (HumanEval / MBPP / LiveCodeBench), where the reward signal comes from behavioral \emph{outcomes} rather than from internal substrate. The three-arc negative here narrows the hypothesis space but does not close it.

\paragraph{For calibration research.} The Gemma~4~E2B baseline disposition finding (assertion asymmetry $-0.009$; $91\%$ assertion rate regardless of correctness on Chef) is a quantified instance of confidence-correctness decoupling on a specific model and domain. Our measurement anchors this to an external gold standard, providing a concrete baseline for future calibration interventions on small instruct-tuned models.

\paragraph{For interpretability and probing.} High in-distribution CV AUC on linear probes of pooled hidden states can collapse to chance on fresh data from the \emph{same} distribution, same model, same judge, and same rubric. This is not new; probe reliability has been flagged before \cite{hewitt2019designing, belinkov2022probing}. However, the effect size here (AUC $0.68 \to 0.52$, asym $-0.34 \to -0.04$) and the fact that a per-axis sweep, multi-layer pooling, and cross-model replication all reproduce the failure is, to our knowledge, the cleanest demonstration of the gap between within-distribution CV and fresh transfer for a frozen-base sidecar theory. The judge-versus-probe gap we surface is a useful quantity that we recommend reporting whenever a probe-based behavioral edit is proposed.

\paragraph{For negative results.} Three failed operator classes on five models, with mechanism, is not ``we couldn't get it to work''; it is evidence for a property of the substrate under the tested conditions. We argue negative results at this granularity are publishable and load-bearing for the field, and we release all artifacts to support independent reproduction.

% =====================================================================
\section{Limitations}
% =====================================================================

\begin{itemize}[leftmargin=*]
\item \textbf{Scale ceiling.} All tested bases are $\leq 2.3$B effective. The failures may not generalize to 4B, 7B, or larger. We have not tested larger bases.
\item \textbf{Domain.} Arc~1 covers coding and French cuisine. Arcs~2--3 use a single judge-based domain (Chef error-correction). Verifiable domains (code with execution oracle, math with checkers) may behave differently under Arc~2--3 operators; we have not tested them.
\item \textbf{Probe family.} All probes in Arc~3 are linear \texttt{LogReg} on pooled hidden states. Non-linear probes (kernel SVM, small MLP), token-level probes, and attention-weighted probes are untested. Our falsification is faithful to the original linear-probe theory the sidecar was specified under, but does not rule out non-linear readouts.
\item \textbf{Alternative reward signals.} Judge-as-reward RLHF (PPO/GRPO with the v4 judge as RM) is a distinct operator class not covered in this paper. We neither endorse nor falsify it.
\item \textbf{Dispositions tested.} Arc~1 trains on the full seven-disposition pipeline. Arc~2 tempers only Humble and Adversarial-self; Deliberate was not attributable on Chef due to rubric saturation, and Curious, Eager, Persistent, and Self-Improving were not tested. Arc~3 covers the four Chef-locked dispositions plus the factual anchor. The three non-Chef dispositions are therefore ruled out only in Arc~1.
\item \textbf{Judge dependence.} Arcs~2--3 rely on judge-based labels and gold-checklist generation in the Chef domain. Although the v4 rubric is response-blind and anchored to canonical claims generated independently of the model under test, conclusions remain partly dependent on the quality and consistency of the external judge.
\end{itemize}

% =====================================================================
\section{Conclusion}
% =====================================================================

We attempted to train behavioral dispositions into small language models (0.6B--2.3B effective parameters) through three distinct operator classes: imitation into weights via SFT/DPO LoRA, inference-time attention-head tempering on $o_{\text{proj}}$, and a training-free frozen-base $\mathbf{h}_{\text{last}}$ confidence-gated sidecar. Our first-pass evaluation of the first operator reported $+33.9$ MCAS and $+15.3$ HumanEval; the paper drafted around those numbers was withheld from release after a second-pass sanity check revealed both results were artifacts. Three subsequent arcs across five instruct models (Qwen3-0.6B, Qwen3-1.7B, Qwen3.5-0.8B, Gemma~4~E2B, and SmolLM2-1.7B-Instruct) systematically falsified all three operator classes under the tested conditions. The cleanest arc (frozen-base $\mathbf{h}_{\text{last}}$) produced a large-effect within-distribution CV pass ($\text{AUC}{=}0.683$, asym $-0.341$) that collapsed to chance on fresh data ($\text{AUC}{=}0.516$), was replicated-as-fail on a second base through a distinct failure mode (capability cliff versus single output distribution), and is therefore a property of the tested scale range on judge-based domains rather than a tuning issue.

At this scale, the residual stream at the final response token encodes stylistic closure, not content truth-value. Judges read the output distribution, which varies with content; linear probes read the closure position, which does not. We release all code, features, and labels to support independent reproduction, and publish the honest falsification pipeline as the protocol that caught both our own initial false positive and the frozen-base sidecar's CV ceiling. We leave outcome-grounded training in verifiable domains as the remaining plausible direction for small-scale behavioral shaping, and treat it as a separate theory to be tested independently of anything in this paper.

% =====================================================================
\section*{Reproducibility and Artifacts}
% =====================================================================

All scripts, result JSONs, hidden-state feature arrays, and per-response judge labels are released at \texttt{github.com/tinmanlabsl/Models} under MIT license. Arc~3 in particular ships: (a)~the five-step falsification pipeline implementation; (b)~the raw hidden-state features (\texttt{*.npz}) and fresh response JSONs for both Gemma~4~E2B and SmolLM2-1.7B; (c)~the 193 fresh Chef prompts with gold checklists; (d)~the v4 judge rubric and step16 baseline samples; (e)~the running research log documenting the full trajectory from initial apparent positive to closure. Reproduction of the Arc~3 result from cached artifacts requires no GPU; reproduction from scratch requires $\sim$30 minutes on a single RTX 4090.

% =====================================================================

\end{document}